\pdfoutput=1

\documentclass[11pt]{article}

\usepackage{naacl2021}
\usepackage{url}
\usepackage{times}
\usepackage{latexsym}
\usepackage[T1]{fontenc}

\usepackage[utf8]{inputenc}

\usepackage{microtype}
\usepackage{times}
\usepackage{url}
\usepackage{latexsym}

\usepackage{enumitem}
\usepackage{amsmath}
\usepackage{graphicx}
\usepackage{cuted}
\usepackage{subfig}
\usepackage{url}
\usepackage{enumitem}
\usepackage{caption}
\usepackage{multicol}
\usepackage{todonotes}
\DeclareCaptionLabelFormat{andtable}{#1~#2  \&  \tablename~\thetable}
\usepackage[utf8]{inputenc}
\title{Proteno: Text Normalization with Limited Data for Fast Deployment in Text to Speech Systems}

\author{Shubhi Tyagi, Antonio Bonafonte, Jaime Lorenzo-Trueba, Javier Latorre\textsuperscript{}{\thanks{\noindent\textsuperscript{} Work done while at Amazon}} \\
         Amazon Alexa AI}

\begin{document}
\maketitle
\begin{abstract}
\vspace{-1mm}
Developing Text Normalization (TN) systems for Text-to-Speech (TTS) on new languages is hard. We propose a novel architecture to facilitate it for multiple languages while using data less than 3\% of the size of the data used by the state of the art results on English. We treat TN as a sequence classification problem and propose a granular tokenization mechanism that enables the system to learn majority of the classes and their normalizations from the training data itself. This is further combined with minimal precoded linguistic knowledge for other classes. We publish the first results on TN for TTS in Spanish and Tamil and also demonstrate that the performance of the approach is comparable with the previous work done on English. All annotated datasets used for experimentation will be released at \url{https://github.com/amazon-research/proteno}.

\end{abstract}
\vspace{-1mm}
\section{Introduction}
\label{intro}
\vspace{-1mm}
Text-to-speech synthesis (TTS) consists of a number of processing steps that control the conversion of input text to output speech. Text normalization (TN) is usually the first step for any TTS system. It is defined as the process of mapping of written text to its spoken form. As per \citet{Taylor2009}, \textit{semiotic class} denotes things like numbers, dates, times, etc. that are written differently from the way they are verbalized. TN is the process of verbalizing instances of such classes. 

Most systems are entirely based on hard coded rules which is neither scalable across languages nor easy to maintain. Many machine learning based techniques have been proposed for TN but they still have heavy dependency on encoded linguistic knowledge or require considerable amount of annotated data making it diffcult to scale.

The contributions of this paper are as follows: \emph{i)} Presenting Proteno,  a novel architecture for TN with a granular tokenization mechanism, which requires minimal language specific rules, curtails unacceptable errors and is transferable to a large extent to multiple languages, \emph{ii)} Establishing an architecture which can be used to benchmark TN baselines for multiple languages with limited annotated data, \emph{iii)} Release of annotated TN datasets for Tamil and Spanish suitable for TTS systems.

As no benchmark datasets or baselines exist for TN for TTS in Spanish and Tamil, we curated datasets for both and evaluated Proteno on them. We also use the best performing system for TN in English and compare its results with previous work.
\vspace{-5mm}
\section{Related Work}
\vspace{-1mm}
In spite of the success of deep learning approaches in other natural language processing tasks, the problem of TN for TTS systems still remains a challenging one \citep{RNNChallenge}. 
Work has been done to solve TN by pure encoder-decoder methods  particularly Recurrent Neural Networks \citep{Sproat2017RNN,GoogleTextNorm}. However, authors have shown that even though such models can perform well overall, occasionally they can make \textit{``unacceptable errors"} like reading ``\$2" as ``two pounds" and thus rendering the system unsuitable for industrial TTS applications.

To curtail such unacceptable errors, previous work based on \textit{semiotic classification} \cite{Sproat2001Classical,Kestrel,GoogleTextNorm}, are encoded with measures like weighted \textit{Finite-state Transducers} (FSTs) introduced by \newcite{sproat_1996}. FSTs revolve around creating a weak covering grammar which encompasses language specific lexical information.
Although such grammars are easier to create as compared to a full blown grammar, they still need prior knowledge of the language and the language specific rules need to be coded in the system \citep{47344}. To completely induce FST from training data, as suggested by \newcite{GoogleTextNorm}, diverse and large amount of data is required. The data should represent all the forms a particular token can appear in a given language. Such requirements for all semiotic classes limit the reproducibility of such models for a new language with limited annotated data. Other language-agnostic approaches \cite{conkie} also need large amounts of data (5M sentences for each language) as parallel corpus and can also result in unacceptable errors.

Our approach curtails such errors by breaking down complex entities like dates into multiple tokens by a granular tokenization mechanism and also by limiting which tokens can be \textit{accepted} into a class. This mechanism, we will see, also enables the system to rely more on data and disambiguate context for normalizations without requiring the knowledge to be specifically coded in the system. 
\vspace{-1mm}
\section{Proposed Approach}
\vspace{-1mm}
The target normalization can be directly predicted from unnormalized text with a seq2seq architecture \cite{Seq2Seq2014} by treating TN as a \textit{machine translation} task \cite{GoogleTextNorm, mansfield-etal-2019-neural} with the previously mentioned limitations. A way to limit the \textit{unacceptable errors} in such systems would be to limit the kind of normalizations the network can generate for a token \cite{Sproat2017RNN}.

On the other hand, solutions based on semiotic classification convert TN into a sequence tagging problem, where each class has associated mechanisms for normalizing the corresponding unnormalized token(s). It produces verbalizations by first suitably tokenizing the input, then classifying the tokens, and then verbalizing each token according to its corresponding class. These approaches often have a complex tokenization mechanism which is not easily transferable across languages and also need all the possible classes to be exhaustively defined manually.

We solve both these problems by a granular tokenization mechanism which extends the concept of \textit{semiotic classification} to a granular level wherein each unique unnormalized token to normalized token mapping can have a class of its own. The majority of the classes and their appropriate normalizations are automatically learnt from data. 

Our classes represent whether a particular token is of a certain type and convert unnormalized tokens into their normalized form. The goal is to manually define the minimum possible set of classes and all the other classes will be automatically learnt from the data. The system learns when each class should be applied. The solution is divided into 4 stages: \emph{i)} Tokenization of unnormalized data, \emph{ii)} Data preparation, \emph{iii)} Classifying unnormalized tokens into correct classes, \emph{iv)} Normalizing tokens using the corresponding class.
\vspace{-1mm}
\subsection{Tokenizer}
\vspace{-1mm}
Typically TN approaches either assume pre-segmented text by the rule-based standard \citep{Kestrel} which identifies multiword sequences as single segment like dates (Jan. 3, 2016) according to pre-defined semiotic classes or train a neural network for tokenization together with a normalization model \cite{GoogleTextNorm}.
Proteno's tokenization on the other hand, has elementary rules and is deterministic. The segmentation is done by splitting the sentences on spaces and then  further splitting the text when there is a change in the unicode class. Eg: after splitting on spaces a token like `C3PO' will be further split into [`C',`3',`PO']. Such tokenization enables the system to accurately split complex entities like dates while eliminating the need for a manually defined complex class for them. The same tokenization mechanism was used for all the languages tested. Hence, it is transferable across a large group of languages which have words separated by spaces. 

\vspace{-1mm}
\subsection{Data Preparation}
\vspace{-1mm}
While collecting training data, first the unnormalized data is tokenized according to the granular tokenization mechanism described above and then each token is annotated with its corresponding normalized form. Thus, we obtain unnormalized token to normalized token mappings. Eg: a date occurrence `1/1/2020' tokenized as [`1',`/',`1',`/',`2020'] is annotated as [`first',`of', `January',`',`twenty twenty']. For such data annotation, linguistic experts are not needed and this can be done by anyone proficient in the target language.

From our experiments, we observe that for TN the diversity in data is more important than the quantity of data. It is better for the model to see different kinds of normalizations. Hence, while collecting the data we try to ensure decent coverage of different semiotic classes by having at least 25\% of tokens which need normalization (i.e. \textit{non-self}).
\vspace{-5mm}
\subsection{Classes}
\vspace{-1mm}
Each class has 2 functions:
\emph{i)} \textit{\textbf{Accepts}}: This function returns a boolean value of whether a token is accepted by the class. Eg: \textit{cardinal} class accepts only numeric tokens, \emph{ii)} \textit{\textbf{Normalize}}: This is a deterministic function that transforms the unnormalized token into its verbalized form

A token can be classified into a class only if it is \textit{accepted} by it. By restricting the classes a token is accepted into, we limit the kind of normalization output that can be generated. This prevents the model from making \textit{unacceptable errors}.
A token can be accepted by multiple classes which can give different normalizations. In such cases, the model is responsible for predicting the appropriate class from the context. If multiple classes give the same normalization for a token, then during inference it doesn't matter which class is chosen.

We have 2 kinds of classes: \emph{i)} \textbf{Pre-defined}: We define limited number of classes ($\sim$10-15) containing basic normalization rules out of which only a small subset ($\sim$5) contain language specific verbalization rules like \textit{cardinal, ordinal} etc. Rules behind the normalization logic for others like \textit{self, sil, digit, roman numerals,} etc. remain similar across many languages, only the surface form of the normalized version changes. Eg: \textit{self} class indicates that the input is to be passed through as it is and it accepts tokens containing only alphabetical  characters. \textit{Sil} is used to represent silence, which is typically associated with punctuation. It accepts only punctuation or other kinds of symbols which should not be verbalized. \textit{Roman numerals} also have language agnostic logic to convert the roman number into number form and pass it down to its corresponding cardinal or ordinal class for generating desired normalization. 
\emph{ii)} \textbf{Auto Generated (AG)}: Apart from pre-defined classes, the model learns automatically generated classes from the data by going through the unnormalized to normalized token mappings in the dataset. If none of the existing classes (pre-coded or AG) can generate the target normalization for a token in the training data, then a class is automatically generated which accepts only the token responsible for its generation. Its normalize function returns the target normalization observed in the annotated data for that token. Eg: if ``12$\rightarrow$December" is observed in the dataset and if none of the existing classes can generate this normalization then a class \textit{``12\_to\_December\_AG"} is created. This class accepts only ``12" and its normalize function returns ``December". If multiple normalizations are observed for an unnormalized token in the dataset which cannot be generated by existing classes then multiple AGs are stored. AGs enable Proteno to learn majority of the normalizations automatically from data.
\vspace{-1mm}
\subsection{Classification \& Normalization}
\vspace{-1mm}
We model TN as a sequence tagging problem where the input is a sequence of unnormalized tokens and the output is the sequence of classes which can generate the normalized text. Before training the classification model we transform the data to get unnormalized token to class mappings. Eg: [`1',`/',`1',`/',`2020'] $\rightarrow$ [ordinal, /\_to\_of\_AG, 1\_to\_January\_AG, sil, year].
We prepare this data by going over the unnormalized token to normalized token mapping for a sentence and identifying which existing classes can give the target normalization. For a token there can be multiple matching classes. Eg: `2' can be correctly normalized by both cardinal and digit classes. In such cases of multiple matching classes we pick the least frequent class to increase the representation of infrequent classes. This compensates for the imbalance present in the proportion of classes in training set. A more optimum approach to handle cases of multiple matching classes will be explored in the future. 

To classify the sequence of unnormalized tokens to their corresponding classes we experimented with 4 classifiers. We first train a first order Conditional Random Fields (CRFs) \cite{CRF} and then  train neural network (NN) based architectures like Bi-LSTMs \cite{LSTM}, BiLSTM-CRFs \cite{BiLSTM-CRF} and Transformers \cite{Transformer}. We used word embeddings from \newcite{fasttext} for NN systems. \emph{i)} \textit{\textbf{CRF}}: The features used for each unnormalized token in the model are - part of speech tag, list of classes which accept the token as an input, next token in sequence, suffix of the token (from length 1-4),  prefix of the token (from length 1-4), is the token in upper case, is the token numeric and is the token capitalized, \emph{ii)} \textit{\textbf{Bi-LSTM \& BiLSTM-CRFs}}: Using word embeddings and list of classes which accept the token as input features, \emph{iii)} \textit{\textbf{Transformer}}: A Transformer with 6 heads with word embeddings as input features.

For each token we renormalize the probabilities predicted over all classes to only the classes which accept the token. Hence, the model is restricted to classify a token only to one of its few accepted classes. If the system is unable to find a suitable class for the given token (i.e. none of the given classes accept that token) then it gives a empty output instead of an incorrect normalization.
\vspace{-1mm}
\subsection{Aligning tokens in order of verbalization}
\vspace{-1mm}
One of the major challenges in automated TN is handling realignment of tokens which may be required between the written and its spoken form. Our method so far assumes monotonic alignment between the written unormalized tokens and their corresponding spoken normalizations. However, this is not always true. For our chosen languages we saw two exceptions: currency and measure units. Eg: \$3.45 $\rightarrow$ \textit{`Three dollars forty five cents'} and m$^2$  $\rightarrow$ \textit{`squared metres'}. Seq2seq models can naturally learn such kind of realignment from training data \cite{Sproat2017RNN}. However, they are susceptible to errors specially for limited amount of training data for specific classes.
%

Thus, to limit errors in such cases we define some rules. Proteno first recognises instances of currency/measure in the text and prevents them from further splitting by the granular tokenizer. The currency/measure classes have the same granular tokenisation logic along with realignment conditions. They further pass the final tokens to their corresponding classes. Thus, an entity like `\$45.18’ is transformed into [`45', `\$', `18', `.'] and passed to classes as 45$\rightarrow$\textit{cardinal}, \$$\rightarrow$\textit{\$\_to\_dollars\_AG}, 18$\rightarrow$\textit{cardinal}, . $\rightarrow$\textit{\_to\_cents\_AG}. 

As all currency symbols have their own AGs automatically generated from the data there will always be a 1:1 mapping between a symbol and its normalized form. As a result this approach eliminates the possibility of an \textit{unacceptable error} like normalizing \$ $\rightarrow$ Pounds. 

Classes like \textbf{\textit{currency}} and \textbf{\textit{measure}} contain rules that are responsible for realignment only and hence require limited knowledge to be transferred across languages.  The normalization is handled by the already learnt or defined classes. Thus, these classes can be skipped or be used as is for any language which has this kind of realignment. 
\vspace{-1mm}
\section{Experiment Protocol}
\vspace{-1mm}
\subsection{Datasets}
\vspace{-1mm}
As the goal of Proteno is to be applicable for multiple languages, we evaluate the system across 3 languages. For experimentation with new languages we chose Spanish and Tamil. Further, we evaluate Proteno on English, to see how it compares against a language which has more evolved TN systems available. There are no benchmarked annotated TN for TTS datasets available for Tamil and Spanish. \emph{i)} \textit{\textbf{Spanish}}: We gathered data from Wikipedia by selecting sentences rich with entities requiring normalization. Due to budget constraints we could collect a dataset of only 135k tokens (5k sentences), \emph{ii)} \textit{\textbf{Tamil}}: We annotate the data sourced from English-Tamil parallel corpus \cite{Eng-Tamil} and Comparable Corpora \cite{Eckart2013}. From these datasets we sampled 500k tokens (30k sentences) with higher preference towards sentences that needed normalization, \emph{iii)} \textit{\textbf{English}}: We used a portion of the annotated data from \citet{RNNChallenge}. First, we run the Proteno tokenizer over the unnormalized section of the dataset and got unnormalized token to normalized token mappings using elementary rules. By doing so, we were able to correctly match only a portion of the dataset due to its different tokenization. And then, from this subset, 300k tokens (24.7k sentences) were randomly sampled to keep the data size comparable to that used for Tamil. This is 1.5\% of the data used by \newcite{Pramanik_2019} which used first 20M tokens and 3\% of data used by \newcite{GoogleTextNorm} which used first 10M tokens. 
\vspace{-1mm}
\subsection{Training \& Evaluation}
\vspace{-1mm}
Train and test data were split by the ratio of 60:40. We keep a higher test set proportion to have a challenging setting for the systems.
Word Error Rate (WER) is used as the evaluation metric for the different classifiers. We use this metric instead of classification accuracy on the classes in order to enable comparison of results from different TN approaches in the future, which may not use the same tokenization mechanism and hence may not have the same classes benchmarked by previous work.

WER is measured as Levenshtein Distance \cite{Levenshtein_SPD66} between the model prediction and the desired normalization.
Hence, lower WER is desirable. We also report classification accuracy to illustrate that classification accuracy does not directly translate into WER. We first evaluate all the classifiers on Spanish and then choose the classifier with lowest WER  for Tamil and English.  
\section{Results} 
\vspace{-1mm}
\subsection{Spanish}
Due to lack of a standard baseline, we compare the performance of Proteno on Spanish with an existing rule based (RB) system. This is the production TN system containing 1.7k lines of regular expressions code which required extensive linguistic knowledge and programming proficiency.

Normalization was required for \textbf{27\%} of tokens in both the training and the test set. \textbf{10} classes were pre-coded with normalization logic: \textit{self, sil, spell, currency, unit, digit, cardinal, ordinal, roman cardinal and roman ordinal} out of which only \textbf{5} had language specific normalization rules (\textit{spell, cardinal\_masculine, cardinal\_feminine, ordinal\_masculine and ordinal\_feminine}). \textbf{279} AGs were generated from this dataset. 
The WER results from different models is given in Table 1. 
\begin{table}[h!]
\label{table:spanishresults}
\begin{center}
\begin{tabular}{{|l|c|c|}}
 \hline \bf  Models & \bf WER(Train) &  \bf WER(Test) \\ \hline
RB System & 2.3 & 2.3\\
CRF &  0.3 & 1.02*\\
BiLSTM & \bf 0.03 & \bf 0.89*\\
BiLSTM-CRF & 0.04 & \bf 0.89*\\
Transformer & 1.2 & 2.3\\
\hline
\end{tabular}
\end{center}
\captionsetup{justification=centering}
\caption{\label{font-table} WER for CRF vs LSTM vs Transformer. Fields in bold are indicative of best model. * signifies statistically significant difference in comparison to RB}
\end{table}
On the test set, all models except Transformers showed statistically significant difference (p$<<$0.01)  in comparison to the RB system. We can attribute the lower performance of Transformers to lack of \textit{accepted classes} as input features.

Although the numbers suggest that the NN models might be overfitting, we were not able to significantly improve them using regularization techniques. Introducing dropout from 0.1-0.3 increased the train WER from 0.03 to 0.04 but did not impact the test WER. Further increase in dropout increased test WER. We also try replacing the cross entropy loss with the \textit{Weighted Categorical Cross Entropy Loss} to avoid the model's bias towards predicting the dominant class (in this case \textit{`self'}). This loss function decreased the train WER from 0.03 to 0.027 but it did not impact the test WER.

For most of the classes CRFs and NN models performed at par with each other. Classification accuracy by the models is given in Table 2. However, low classification accuracy, though indicative of inaccurate normalization, does not directly translate into higher WERs. Multiple classes can give the same normalization and thus there is no unique correct class. This is particularly prevalent in some cases of number instances where \textit{cardinal\_masculine} and \textit{cardinal\_feminine} can be used interchangeably.

Even though Transformers give unstable performance in class prediction, they still give a low enough WER. This particular iteration has a bias towards predicting \textit{cardinal\_ masculine} over \textit{cardinal\_ feminine}. This bias changes with different iterations but the WER remains consistent as the normalization output remains unaffected. 
\begin{table*}[t!]
\label{table:spanishclassification}
\begin{center}
\small
\scalebox{1.0}{
\begin{tabular}{| *{11}{c|}}
\hline  & \multicolumn{2}{c|} {\bf Token Proportion} & \multicolumn{2}{c|} {\bf CRF} & \multicolumn{2}{c|}{\bf  BiLSTM} & \multicolumn{2}{c|}{\bf BiLSTM-CRF} & \multicolumn{2}{c|}{\bf Transformers}  \\ \hline
  & Train & Test & Train & Test & Train & Test & Train & Test & Train & Test  \\ \hline
\bf Accuracy & & &  99.7  & \bf 99.1 & \bf 99.9 &  99.01 & 99.99 &  98.9 & 93.0 & 92.8  \\ \hline
\bf Accuracy per class  & & & & & & & & & &\\
	`self' & 70.5 & 70 & \bf 100 & \bf 100 &  \bf 100 &  99.9 & \bf 100 &  99.9 & \bf 100 & 99.8 \\	
	`sil' & 13.24 & 13 & 99.7 & \bf 99.8 &  \bf 100 &  99.5 & 99.99 & 99.6 & \bf 100 & 98.7\\
	 \bf Others & & &  98.0 &  93.2  & \bf  99.9 &  93.06 & \bf 99.9 & \bf 95.9 & 44.2 & 48.3 \\	
	`es\_num\_by\_num\_cardinal' & 2.14 & 2.1 & \bf 99.9 & \bf 99.2 & \bf 99.9 &  \bf 99.2 & \bf 99.9 & 98.6 & 3.85 & 2.4 \\
	`es\_cardinal\_feminine' & 3.8 & 3.8 & 98.9 & \bf 96.7 &  \bf 100 &  93.5 & 99.9 & 92.8 & 37.7 & 41.6 \\
	`es\_ordinal\_masculine' & 0.38 & 0.4 & 95.2 & 96.7 &  99.7  &  96.7 & \bf 100 & \bf 97.1 & 0 & 1.9 \\
	`spell' & 0.62 & 0.57 & 98.7 & 96.0  &  \bf 100 &  75.2 & \bf 100 & 71.1 & 99.6 & \bf 99.3 \\	
	`es\_cardinal\_masculine' & 1.75 & 2.16 &  98.2 &  89.2 &  \bf 100 &  \bf 98.8 & 99.8 & 98.6 & 87.0 & 88.1\\
			`es\_ordinal\_feminine' & 0 & 0.00004 &  n/a  &  0.0 & n/a &  0  & n/a & 0  & n/a & \bf 100 \\ 
	`mean'  &  7.63 & 8 &  97.6 & \bf 92.6 & \bf 99.9 &  89.7 & \bf 99.9 & 88.5 & 47.9 & 51.9  \\ \hline
\end{tabular}
}
\end{center}
\captionsetup{justification=centering}
\caption{\label{font-table} Token proportions and classification accuracy across systems for Spanish. `mean' depicts the average accuracy of the remaining  precoded and all the AG classes. Bold font highlights the best results }
\end{table*}
\subsection{Tamil}
For Tamil, we have \textbf{8} pre-coded classes  \textit{(self\_english, self\_tamil, sil, spell, currency, digit, cardinal and ordinal)} out of which only \textbf{3} are encoded with language specific normalization logic (\textit{cardinal, ordinal and spell}) and \textbf{74} AGs were generated from the dataset. To normalize text on Tamil corpus, we trained the system which performed the best on Spanish i.e. BiLSTMs with the same configurations. The model gave a WER=\textbf{0.6} on the train set and WER=\textbf{3.3} on the test set. The token proportion and high level classification accuracy results for the tokens are detailed in Table 3.
\subsection{English}
To evaluate the potential of the approach and benchmark it with existing work we trained Proteno on English. The model had \textbf{8} precoded classes \textit{(self, sil, spell, cardinal, ordinal, digit, roman, units, year)} out of which only \textbf{4} classes contained language specific rules (\textit{spell, cardinal, ordinal, year}). \textbf{2658} AGs were generated from the data. The number of AGs in English are significantly higher than the ones generated for Tamil or Spanish as English tends to use much more abbreviations in written form as compared to the other two languages. The model achieved a WER=\textbf{0.47} on the train set and a WER=\textbf{2.6} on the test set. High level classification accuracies are detailed in Table 3. Out of the 99.26\%  correctly normalized tokens, 88.2\% of the non-self tokens were normalized via AGs i.e. \textbf{88.2\%} of the normalizations were learnt automatically from data without relying on pre-coded linguistic knowledge. 
\begin{table*}[t!]
\label{table:engtamilresults}
\begin{center}
\scalebox{0.8}{
\begin{tabular}{{|l|c|c|c|c|c|c|c|c|c|c|}}
 \hline  \bf Language &  \multicolumn{2}{c|} {\bf Proportion of}  & \multicolumn{2}{c|}{ \bf Proportion of} & \multicolumn{2}{c|} {\bf Accuracy on} & \multicolumn{2}{c|} {\bf Accuracy on} & \multicolumn{2}{c|}{\bf Overall} \\ 
& \multicolumn{2}{c|}{\bf self tokens} & \multicolumn{2}{c|}{\bf other tokens} & \multicolumn{2}{c|}{\bf  self tokens} & \multicolumn{2}{c|}{\bf other tokens} &  \multicolumn{2}{c|}{\bf Accuracy} \\ \hline
& Train & Test & Train & Test & Train & Test & Train & Test & Train & Test \\ \hline
Tamil  & 0.73  &  0.75 & 0.27 & 0.25 &  99.99 & 99.99 & 99.94 & 96.49 & 99.98 & 99.12 \\
English & 0.72 &  0.71 & 0.28 & 0.29 &  99.97 & 99.99 & 99.55 & 97.5 & 99.85 & 99.26  \\
\hline
\end{tabular}
}
\end{center}
\captionsetup{justification=centering}
\caption{\label{font-table} Token proportions and classification accuracy for Tamil and English}
\end{table*}

\begin{table*}[t!]
\centering
\scalebox{0.75}{
\begin{tabular}{{|c|c|c|c|c|c|c|c|c|c|c|c|}}
\hline & \bf Plain & \bf Punct & \bf Date & \bf Cardinal & \bf Verbatim & \bf Measure & \bf Ordinal & \bf Decimal & \bf Digit & \bf Fraction & \bf Letters \\ \hline
\bf Train Proportion & 70.2 & 18.8 & 6.13 & 1.13 & 0.82 & 0.21 &  0.11 & 0.20 &  0.04 & 0.0 & 2.39 \\
\bf  Test Proportion  & 70.3 & 18.7 & 6.08 &  1.30 & 0.71 & 0.19 & 0.15 & 0.16 & 0.04 & 0.001 & 2.27 \\
\bf Proteno & 99.9 & 100 & 98.16 &  99.08 & 96.97 & 96.09 & 73.05 & 90.0 & 41.30 & 100.0 &  79.18 \\
\bf P\&H & 99.4 & 99.9 &  99.7 & 99.4 & 99.4 & 97.1 & 98.0 & 98.9 & 79.5 & 92.3 & 97.1 \\
\bf Z & 99.9 & 99.9 &  99.5 & 99.4 & 99.9 & 97.2 & 98.1 & 100 &  86.4 & 81.3 & 97.5 \\ \hline
\end{tabular}
}
\captionsetup{justification=centering, font=normalsize}
\caption{\label{font-table} English Classification Accuracy: Proteno vs \newcite{Pramanik_2019} vs \newcite{GoogleTextNorm}}
\label{tab:GoogleCats}
\end{table*}
It is not possible to directly compare our results with previous work done on English TN \cite{Pramanik_2019,GoogleTextNorm} as these works report classification accuracy on 16 manually defined classes and not WER. Moreover, Proteno does not have the same set of classes due to its granular tokenization mechanism. It also uses only 1.5\%-3\% of the dataset used by them and further splits it into train and test set. It cannot use the full dataset due to differing tokenization mechanisms which result into mismatch in the alignment between the unnormalized token and their corresponding normalized forms. However, we extract their pre-defined categories on the dataset we used and evaluate how many tokens within them were normalized correctly. In Table 4 we compare Proteno accuracy with the accuracy reported by \newcite{Pramanik_2019} (P\&H) and by \newcite{GoogleTextNorm} (Z). It illustrates the token normalization accuracy achieved by Proteno on the test dataset for all the categories which had instances in the small subset we have used.

Proteno performs at par with the other systems for most of the categories in spite of seeing much fewer instances in the train set. For complex entities likes \textit{date} Proteno gave 98.16\% accuracy on the 6\% tokens available in test set. The system (Z) gives 99.5\% accuracy on its set by using a covering grammar learnt from large amounts of data. We observe comparable performance for another complex category like \textit{measure}. On the other hand, we see a significant drop in Proteno's performance when normalizing \textit{ordinal} and \textit{digit}. This is due to low representation of these classes during training and hence during inference the model has a bias towards predicting \textit{cardinal} over them when seen in similar context. This bias can be addressed by having a more equitable representation of instances of \textit{cardinals, ordinals} and \textit{digits} during training.
\vspace{-2mm}
\section{Conclusions}
\vspace{-2mm}
We propose a novel architecture suitable for scaling Text Normalization for TTS across languages using minimal language specific rules, limited annotated dataset and while curbing \textit{unacceptable errors} which makes it suitable for fast deployment in industry applications. We treat Text Normalization as a sequence classification problem while proposing a granular tokenizer which enables majority of normalizations to be automatically learnt from data. We experiment across 3 languages: Spanish, Tamil and English, while pre-coding maximum \textbf{5} classes with language specific logic.  We also demonstrate that datasets of the order of 135k-500k tokens can give competitive performance while still being of a size practical for hand annotation.

Proteno consists of \emph{i)} a granular tokenizer based on Unicode classes, \emph{ii)} a classifier of tokens into classes, either predefined or added based on the tokenized data, and \emph{iii)} the class verbalizers, either defined by linguists for predefined classes or automatically learnt from the data. BiLSTMs give the best performance with WER=\textbf{0.89} for Spanish,  WER=\textbf{3.3} for Tamil and WER=\textbf{2.6} for English. In English, \textbf{88.2\%} of the normalizations were learnt automatically from data while using less than 3\% of the data used in previous work \citep{GoogleTextNorm, Pramanik_2019} and still showed comparable performance.

Given the simplicity of this architecture, we believe that Proteno can be used to benchmark TN for many languages with limited annotated data. However, languages which are not separated by space or highly inflected languages will be a challenge for the proposed system \citep{IcelandicTN}. We leave the adaptation of Proteno to more challenging languages for future work. 

\section*{Acknowledgements}
We would like to thank Denys Savin, Yvonne Flory, Tarek Badr and Anton Nguyen for their foundational contributions to the project and developing the production pipeline.
\bibliography{anthology,custom}

\begin{thebibliography}{21}
\expandafter\ifx\csname natexlab\endcsname\relax\def\natexlab#1{#1}\fi

\bibitem[{{A. {Conkie} and A. {Finch}}({2020})}]{conkie}
{A. {Conkie} and A. {Finch}}. {2020}.
\newblock {Scalable Multilingual Frontend for TTS}.
\newblock In \emph{{ICASSP 2020 - 2020 IEEE International Conference on
  Acoustics, Speech and Signal Processing (ICASSP)}}, pages 6684--6688.

\bibitem[{Ebden and Sproat(2014)}]{Kestrel}
Peter Ebden and Richard Sproat. 2014.
\newblock \href {https://doi.org/10.1017/S1351324914000175} {{The Kestrel TTS
  text normalization system}}.
\newblock \emph{Natural Language Engineering}, 21:333--353.

\bibitem[{Eckart and Quasthoff(2013)}]{Eckart2013}
Thomas Eckart and Uwe Quasthoff. 2013.
\newblock \href {https://doi.org/10.1007/978-3-642-20128-8_8}
  {\emph{{Statistical Corpus and Language Comparison on Comparable Corpora}}},
  pages 151--165. Springer Berlin Heidelberg, Berlin, Heidelberg.

\bibitem[{Hochreiter and Schmidhuber(1997)}]{LSTM}
Sepp Hochreiter and J\"{u}rgen Schmidhuber. 1997.
\newblock \href {https://doi.org/10.1162/neco.1997.9.8.1735} {{Long Short-Term
  Memory}}.
\newblock \emph{Neural Comput.}, 9(8):1735--1780.

\bibitem[{Huang et~al.(2015)Huang, Xu, and Yu}]{BiLSTM-CRF}
Zhiheng Huang, Wei Xu, and Kai Yu. 2015.
\newblock \href {http://arxiv.org/abs/1508.01991} {{Bidirectional {LSTM-CRF}
  Models for Sequence Tagging}}.
\newblock \emph{CoRR}, abs/1508.01991.

\bibitem[{Lafferty et~al.(2001)Lafferty, McCallum, and Pereira}]{CRF}
John~D. Lafferty, Andrew McCallum, and Fernando C.~N. Pereira. 2001.
\newblock \href {http://dl.acm.org/citation.cfm?id=645530.655813} {{Conditional
  Random Fields: Probabilistic Models for Segmenting and Labeling Sequence
  Data}}.
\newblock In \emph{Proceedings of the Eighteenth International Conference on
  Machine Learning}, ICML '01, pages 282--289, San Francisco, CA, USA. Morgan
  Kaufmann Publishers Inc.

\bibitem[{Levenshtein(1966)}]{Levenshtein_SPD66}
Vladimir~Iosifovich Levenshtein. 1966.
\newblock Binary codes capable of correcting deletions, insertions and
  reversals.
\newblock \emph{Soviet Physics Doklady}, 10(8):707--710.
\newblock Doklady Akademii Nauk SSSR, V163 No4 845-848 1965.

\bibitem[{Mansfield et~al.(2019)Mansfield, Sun, Liu, Gandhe, and
  Hoffmeister}]{mansfield-etal-2019-neural}
Courtney Mansfield, Ming Sun, Yuzong Liu, Ankur Gandhe, and Bj{\"o}rn
  Hoffmeister. 2019.
\newblock \href {https://doi.org/10.18653/v1/N19-2024} {Neural text
  normalization with subword units}.
\newblock In \emph{Proceedings of the 2019 Conference of the North {A}merican
  Chapter of the Association for Computational Linguistics: Human Language
  Technologies, Volume 2 (Industry Papers)}, pages 190--196, Minneapolis,
  Minnesota. Association for Computational Linguistics.

\bibitem[{Mikolov et~al.(2018)Mikolov, Grave, Bojanowski, Puhrsch, and
  Joulin}]{fasttext}
Tomas Mikolov, Edouard Grave, Piotr Bojanowski, Christian Puhrsch, and Armand
  Joulin. 2018.
\newblock {Advances in Pre-Training Distributed Word Representations}.
\newblock In \emph{Proceedings of the International Conference on Language
  Resources and Evaluation (LREC 2018)}.

\bibitem[{Nikulásdóttir and Guðnason(2019)}]{IcelandicTN}
Anna~Björk Nikulásdóttir and Jón Guðnason. 2019.
\newblock \href {https://doi.org/10.21437/Interspeech.2019-2367}
  {{Bootstrapping a Text Normalization System for an Inflected Language.
  Numbers as a Test Case}}.
\newblock In \emph{Proc. Interspeech 2019}, pages 4455--4459.

\bibitem[{Pramanik and Hussain(2019)}]{Pramanik_2019}
Subhojeet Pramanik and Aman Hussain. 2019.
\newblock \href {https://doi.org/10.1016/j.specom.2019.02.003} {{Text
  normalization using memory augmented neural networks}}.
\newblock \emph{Speech Communication}, 109:15–23.

\bibitem[{Ramasamy et~al.(2012)Ramasamy, Bojar, and
  {\v{Z}}abokrtsk{\'{y}}}]{Eng-Tamil}
Loganathan Ramasamy, Ond{\v{r}}ej Bojar, and Zden{\v{e}}k
  {\v{Z}}abokrtsk{\'{y}}. 2012.
\newblock {Morphological Processing for English-Tamil Statistical Machine
  Translation}.
\newblock In \emph{Proceedings of the Workshop on Machine Translation and
  Parsing in Indian Languages ({MTPIL}-2012)}, pages 113--122.

\bibitem[{Sodimana et~al.(2018)Sodimana, Silva, Sproat, Theeraphol, Li, Gutkin,
  Sarin, and Pipatsrisawat}]{47344}
Keshan Sodimana, Pasindu~De Silva, Richard Sproat, A~Theeraphol, Chen~Fang Li,
  Alexander Gutkin, Supheakmungkol Sarin, and Knot Pipatsrisawat. 2018.
\newblock \href
  {https://www.isca-speech.org/archive/SLTU_2018/pdfs/Keshan2.pdf} {Text
  normalization for bangla, khmer, nepali, javanese, sinhala, and sundanese tts
  systems}.
\newblock In \emph{6th International Workshop on Spoken Language Technologies
  for Under-Resourced Languages (SLTU-2018)}, pages 147--151, 29--31 August,
  Gurugram, India.

\bibitem[{Sproat(1996)}]{sproat_1996}
Richard Sproat. 1996.
\newblock \href {https://doi.org/10.1017/S1351324997001654} {{Multilingual text
  analysis for text-to-speech synthesis}}.
\newblock \emph{Natural Language Engineering}, 2(4):369–380.

\bibitem[{Sproat et~al.(2001)Sproat, Black, Chen, Kumar, Ostendorf, and
  Richards}]{Sproat2001Classical}
Richard Sproat, Alan~W. Black, Stanley Chen, Shankar Kumar, Mari Ostendorf, and
  Christopher Richards. 2001.
\newblock \href {https://doi.org/10.1006/csla.2001.0169} {{Normalization of
  Non-standard Words}}.
\newblock \emph{Comput. Speech Lang.}, 15(3):287--333.

\bibitem[{Sproat and Jaitly(2016)}]{RNNChallenge}
Richard Sproat and Navdeep Jaitly. 2016.
\newblock \href {http://arxiv.org/abs/1611.00068} {{{RNN} Approaches to Text
  Normalization: {A} Challenge}}.
\newblock \emph{CoRR}, abs/1611.00068.

\bibitem[{Sproat and Jaitly(2017)}]{Sproat2017RNN}
Richard Sproat and Navdeep Jaitly. 2017.
\newblock \href {https://doi.org/10.21437/Interspeech.2017-35} {{An RNN Model
  of Text Normalization}}.
\newblock In \emph{Proc. Interspeech 2017}, pages 754--758.

\bibitem[{Sutskever et~al.(2014)Sutskever, Vinyals, and Le}]{Seq2Seq2014}
Ilya Sutskever, Oriol Vinyals, and Quoc~V. Le. 2014.
\newblock \href {http://arxiv.org/abs/1409.3215} {{Sequence to Sequence
  Learning with Neural Networks}}.
\newblock \emph{CoRR}, abs/1409.3215.

\bibitem[{Taylor(2009)}]{Taylor2009}
Paul Taylor. 2009.
\newblock \href {https://doi.org/10.1017/CBO9780511816338}
  {\emph{{Text-to-Speech Synthesis}}}.
\newblock Cambridge University Press.

\bibitem[{Vaswani et~al.(2017)Vaswani, Shazeer, Parmar, Uszkoreit, Jones,
  Gomez, Kaiser, and Polosukhin}]{Transformer}
Ashish Vaswani, Noam Shazeer, Niki Parmar, Jakob Uszkoreit, Llion Jones,
  Aidan~N. Gomez, Lukasz Kaiser, and Illia Polosukhin. 2017.
\newblock \href {http://arxiv.org/abs/1706.03762} {{Attention Is All You
  Need}}.
\newblock \emph{CoRR}, abs/1706.03762.

\bibitem[{Zhang et~al.(2019)Zhang, Sproat, H.~Ng, Stahlberg, Peng, Gorman, and
  Roark}]{GoogleTextNorm}
Hao Zhang, Richard Sproat, Axel H.~Ng, Felix Stahlberg, Xiaochang Peng, Kyle
  Gorman, and Brian Roark. 2019.
\newblock \href {https://doi.org/10.1162/COLI_a_00349} {{Neural Models of Text
  Normalization for Speech Applications}}.
\newblock \emph{Computational Linguistics}, pages 1--49.

\end{thebibliography}
\bibliographystyle{acl_natbib}
\end{document}